\documentclass[letterpaper, 10 pt, conference]{ieeeconf}  %

\IEEEoverridecommandlockouts                              %

\usepackage{graphics} %
\usepackage{amsmath} %
\usepackage{amssymb}  %
\usepackage{verbatim}
\usepackage{xcolor}
\usepackage{mathtools}
\usepackage{booktabs}
\usepackage{myformat}
\usepackage{nicefrac}
\usepackage[sort]{cite}
\usepackage[keeplastbox]{flushend}

\usepackage[caption=false]{subfig}

\usepackage{tikz}
\usepackage{pgfplots}
\pgfplotsset{compat = 1.6}
\usetikzlibrary{calc, shapes.callouts, shapes.arrows}
\usetikzlibrary{external}
\newif\ifblacktext
\blacktextfalse

\newcommand{\evcam}{event-camera}
\newcommand{\Evcam}{Event-camera}
\newcommand{\evcams}{\evcam s}
\newcommand{\Evcams}{\Evcam s}

\ifblacktext
\newcommand{\todo}[1]{#1}
\else
\newcommand{\todo}[1]{}
\fi

\newcommand\MYhyperrefoptions{bookmarks=true,bookmarksnumbered=true,%
pdfpagemode={UseOutlines},plainpages=false,pdfpagelabels=true,%
colorlinks=true,citecolor={black},%
pdftitle={Event-Based Angular Velocity Regression with Spiking Networks},%
pdfsubject={Spiking Neural Networks, Neuromorphic Engineering, Computer Vision},%
pdfauthor={M. Gehrig, S. Bam Shrestha, D. Mouritzen, D. Scaramuzza},%
pdfkeywords={Spiking Neural Networks, Event Cameras}}%
\usepackage[\MYhyperrefoptions,pdftex]{hyperref}

\definecolor{somegray}{rgb}{0.5, 0.5, 0.5}
\newcommand{\darkgrayed}[1]{\textcolor{somegray}{#1}}
\makeatletter
\newcommand*\titleheader[1]{\gdef\@titleheader{#1}}
\AtBeginDocument{%
  \let\st@red@title\@title
  \def\@title{%
    \vskip-3em
    \bgroup\normalfont\large\centering\@titleheader\par\egroup
    \vskip1.5em\st@red@title}
}
\makeatother

\titleheader{\darkgrayed{This paper has been accepted for publication at the\\
IEEE International Conference on Robotics and Automation (ICRA), Paris, 2020.
\copyright IEEE}}

\title{\LARGE \bf
Event-Based Angular Velocity Regression with Spiking Networks
}

\author{Mathias Gehrig$^{1}$, Sumit Bam Shrestha$^{2}$, Daniel Mouritzen$^{1}$, and Davide Scaramuzza$^{1}$%
\thanks{$^{1}$Mathias Gehrig, Daniel Mouritzen and Davide Scaramuzza are with the Robotics and Perception Group, Dep. of Informatics, University of Zurich, and Dep. of Neuroinformatics, University of Zurich and ETH Zurich, Switzerland--- \url{http://rpg.ifi.uzh.ch.}
Their work was supported by the SNSF-ERC Starting Grant and the Swiss National Science Foundation through the National Center of Competence in Research (NCCR) Robotics.}%
\thanks{$^{2}$Sumit Bam Shrestha is with Temasek Laboratories, National University of Singapore, Singapore. His work is partially supported by Programmatic grant no. A1687b0033 from the Singapore government’s Research, Innovation and Enterprise 2020 plan (Advanced Manufacturing and Engineering domain)
}%
}

\begin{document}
\bstctlcite{IEEEexample:BSTcontrol}

\maketitle
\thispagestyle{empty}
\pagestyle{empty}

\begin{abstract}
Spiking Neural Networks (SNNs) are bio-inspired networks that process information conveyed as temporal spikes rather than numeric values. An example of a sensor providing such data is the \evcam. It only produces an event when a pixel reports a significant brightness change. Similarly, the spiking neuron of an SNN only produces a spike whenever a significant number of spikes occur within a short period of time. Due to their spike-based computational model, SNNs can process output from event-based, asynchronous sensors without any pre-processing at extremely lower power unlike standard artificial neural networks. This is possible due to specialized neuromorphic hardware that implements the  highly-parallelizable concept of SNNs in silicon. Yet, SNNs have not enjoyed the same rise of popularity as artificial neural networks. This not only stems from the fact that their input format is rather unconventional but also due to the challenges in training spiking networks. Despite their temporal nature and recent algorithmic advances, they have been mostly evaluated on classification problems. We propose, for the first time, a temporal regression problem of numerical values %
given events from an \evcam. We specifically investigate the prediction of the 3-DOF angular velocity of a rotating \evcam~with an SNN.
The difficulty of this problem arises from the prediction of angular velocities continuously in time directly from irregular, asynchronous event-based input. Directly utilising the output of \evcams~without any pre-processing ensures that we inherit all the benefits that they provide over conventional cameras. That is high-temporal resolution, high-dynamic range and no motion blur. To assess the performance of SNNs on this task, we introduce a synthetic \evcam~dataset generated from real-world panoramic images and show that we can successfully train an SNN to perform angular velocity regression.
\todo{Add link to open source code and dataset when done}

\end{abstract}
\section*{Supplementary Matieral}
Code is available at\\\url{https://tinyurl.com/snn-ang-vel}
\section{Introduction}\label{sec:introduction}

\begin{figure}[t]
    \centering
    \includegraphics{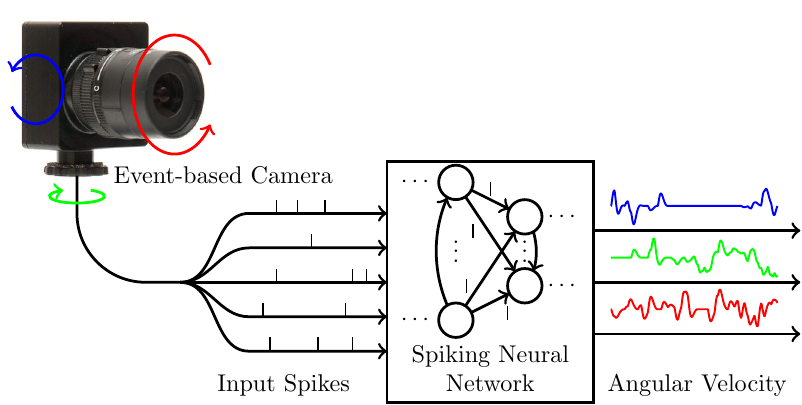}
    \caption{Processing pipeline for event-based angular velocity regression using a spiking neural network.}
    \label{fig:systemDiagram}
\end{figure}

A spiking neural network (SNN) is a bio-inspired model consisting of spiking neurons as the computational model. A spiking neuron is a mathematical abstraction of a biological neuron, which processes temporal events called spikes and also outputs spikes \cite{Gerstner2002}. It has a one-dimensional internal state (potential), that is governed by first-order dynamics. Whenever a spike arrives, the potential gets excited but decays again if no other spikes are registered close in time. In case of the potential reaching a certain threshold, a spiking neuron emits a spike to connected neurons and resets its own potential. If we now link many neurons together we create a dynamical neural network that processes information with spikes rather than numeric values. %
This crucial difference is why SNNs and artificial neural networks (ANNs) are not necessarily competitors but rather models that are intrinsically suitable for a distinct set of problems. As an example, SNNs are able to process asynchronous, irregular data from event-based cameras directly \cite{cohen2016skimming}, without pre-processing events \cite{Gehrig19arxiv} and at extremely low power \cite{Moradi18tbcas}. We refer to the survey paper by Gallego et al. \cite{Gallego19arxiv} for an introduction to event-based vision.

Even training feedforward spiking neural networks is notoriously difficult. The main reason for this is that the spike-generation mechanism within a spiking neuron is non-differentiable. Furthermore, spikes have a temporal effect on the dynamics of the receiving neuron and introduce a temporal dimension to the error assignment problem. As a result, standard backpropagation\cite{rumelhart2002learning} is not directly applicable to SNNs. Nonethelesss, the majority of research on supervised learning for SNNs has taken inspiration from backpropagation to solve the error assignment problem. However, some algorithms are only designed for a single neuron\cite{Ponulak2005, MOHEMMED2012, GuetigSompolinsky2006}, ignore the temporal effects of spikes\cite{Lee2016, Jin2018, Neftci2019, Zenke2018} or employ heuristics for successful learning\cite{Bohte2002a, Booij2005, ShresthaSong2016b, Shrestha2018}  on small-scale problems.
Although SNNs are a natural fit for spatio-temporal problems, they have largely been applied to classification problems \cite{shrestha2018slayer, Jin2018, Wu2018, Neftci2019, Zenke2018, TavanaeiGhodratiKheradpishehEtAl2019, Esser2016, Diehl2015, Rueckauer2017}, except for a few demonstrations addressing learning of spike sequences (spike-trains)~\cite{Zenke2018, shrestha2018slayer, Shrestha2016}. Therefore, it is unclear which algorithm can successfully train multi-layer architectures for tasks beyond classification.

\subsection{Contributions}
In this work, we explore the utility of SNNs to perform regression of numeric values in the continuous-time domain from event-based data. %
To the best of our knowledge, this problem setting has not been explored in SNN literature at the time of the submission. %
The framework is illustrated in figure~\ref{fig:systemDiagram}. The task of our choice is angular velocity (tilt, pan, roll rates) regression of a rotating \evcam. Successful attempts to this task require a training algorithm that is able to perform accurate spatio-temporal error assignment. This might not be necessary for performing classification on neuromorphic datasets and, thus, raises a challenge for SNNs.

Our problem setting offers the context to approach a number of unanswered questions:
\begin{itemize}
    \item How do we formulate a continuous-time, numeric regression problem for SNNs?
    \item Can current state-of-the-art SNN-based learning approaches solve temporal problems beyond classification?
    \item What kind of architecture performs well on this task?
    \item Can SNNs match the performance of ANNs in numeric regression tasks?
\end{itemize}
As a first building block, we introduce a large-scale synthetic dataset from real-world scenes using a state-of-the-art \evcam~simulator \cite{Rebecq18corl}. This dataset provides precise ground truth for angular velocity which is used both for training and evaluation of the SNN. We use this dataset to successfully train a feedforward convolutional SNN architecture that predicts tilt, pan, and roll rates at all times with a recently proposed supervised-learning approach \cite{shrestha2018slayer}. In addition to that, we show that our network predicts accurately at the full range of angular velocities and extensively compare against ANN baselines designed to perform this task in discrete-time.

In summary, our contributions are:
\begin{itemize}
    \item The introduction of a continuous-time regression problem for spiking neural networks along with a dataset for reproducibility.
    \item A novel convolutional SNN architecture designed for regression of numeric values.
    \item A detailed evaluation against state-of-the-art ANN models crafted for event-based vision problems.
\end{itemize}
\section{Related Work}\label{sec:relwork}
Currently, artificial neural networks are the de facto computational model of choice for a wide range problems, such as classification, time series prediction, regression analysis, sequential decision making etc. Spiking neural networks add additional biological relevance in these architectures with the use of a spiking neuron as the distributed computational unit. With the promise of increased computational ability\cite{Maass1996, Maass1996a} and low power computation using neuromorphic hardware\cite{Davies2018, Merolla2014, Neckar2019, Furber2014}, SNNs show their potential as computational engines, especially for processing event-based data from neuromorphic sensors\cite{Lichtsteiner05rme, chan2007aer}.

One of the major bottlenecks in realizing the computational potential of SNNs has been the fact that backpropagation is not directly applicable to training SNNs. This is due to the non-differentiable nature of the spike generation mechanism in spiking neurons. Nevertheless, there have been some efforts in tailoring backpropagation for SNNs. Prominent examples are event-based backpropagation methods \cite{Bohte2002a,Shrestha2016, schrauwen2004improving} that backpropagate error at spike-times. However, they have shown limited success. In recent times, the idea of using a continuous function as a proxy for spike function derivative has been used effectively\cite{shrestha2018slayer, Wu2018, Neftci2019, Zenke2018} for relatively deep feedforward SNNs. \cite{shrestha2018slayer, Wu2018} also take into account the temporal dynamics present in SNNs to assign error in time. Still, gradients can only be computed approximately with these methods. As a result, it is unclear whether there are better performing algorithms for supervised learning for feedforward SNNs.

Almost all of the reported use cases of SNNs in the aforementioned methods are classification problems, such as image classification~\cite{Lecun1998, Krizhevsky2009}, neuromorphic classification~\cite{Orchard2015}, action recognition~\cite{Amir2017}, etc. When it comes to regression problems, there are demonstrations of toy spike-to-spike translation problems \cite{Zenke2018, shrestha2018slayer} for which a target spike-train is learned. To the best of our knowledge, there is currently no published work exploring the use of SNNs for predicting numeric values in continuous-time.

\todo{Mention Guillermos work on angular velocity tracking here or in the methodology}
\todo{Mention spiking yolo}

\section{Methodology}\label{sec:method}

\subsection{Spiking Neurons}
\begin{figure}
    \centering
    \includegraphics[width=\linewidth]{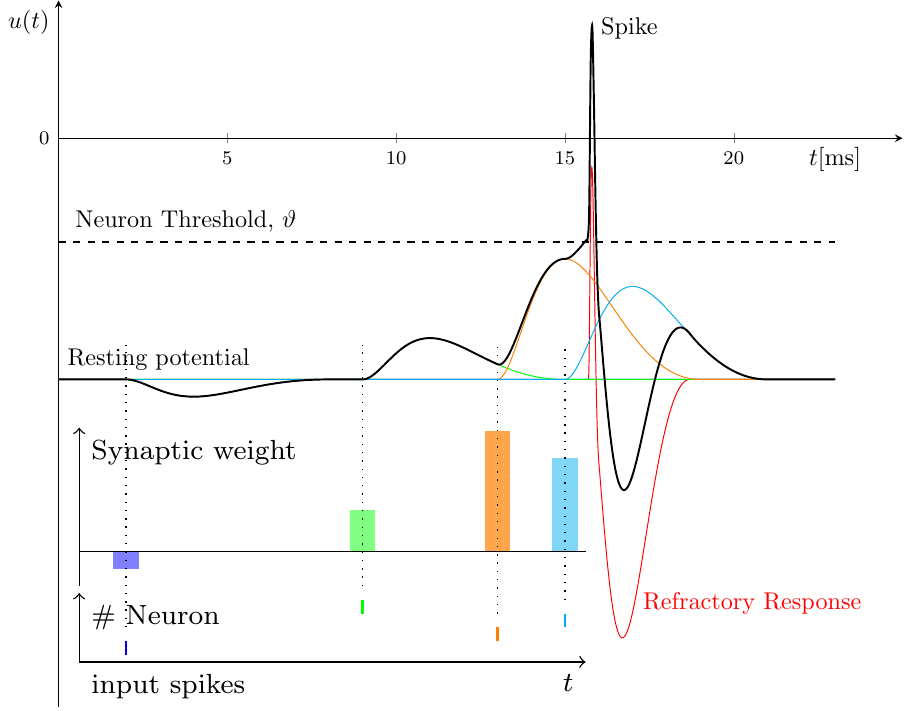}
    \caption{Dynamics of a spiking neuron. A spiking neuron is exciting by incoming spikes according to the corresponding synaptic weights. If the potential reaches the neuron threshold, a spike is emitted and the potential of the neuron is reset by a refractory response. \todo{remove potential overshoot when spiking to avoid confusion}}
    \label{fig:neuronDynamics}
\end{figure}

Spiking neurons model the dynamics of a biological neuron. They receive spikes, which are short pulses of voltage surge, and distribute their effects in time to form a post-synaptic potential (PSP). The magnitude and sign of the PSP is determined by the synaptic weight corresponding to the spike. Finally, the accumulation of all the PSPs in a neuron constitutes the sub-threshold membrane potential $u(t)$. This process is illustrated in figure \ref{fig:neuronDynamics} via spikes from multiple synapses. When the sub-threshold membrane potential is strong enough to exceed the neuron threshold $\vartheta$ the spiking neuron responds with a spike. Immediately after the spike, the neuron tries to suppress its membrane potential so that the spiking activity is regulated. This self-suppression mechanism is called refractory response.

There are various mathematical models in neuroscience that describe the dynamics of a spiking neuron with varying degree of detail: from the complex Hodgkin-Huxley neuron\cite{Hodgkin1952} to the simple Leaky Integrate and Fire neuron\cite{Paugam-Moisy2011, Gerstner2002}. In this paper, we use the Spike Response Model~(SRM)\cite{Gerstner1995}. In SRM, the PSP response is a decoupled, normalized spike response kernel, $\varepsilon(t)$, scaled by the synaptic weight. Similarly, the refractory response is described by a refractory kernel, $\nu(t)$. The SRM is simple, yet versatile enough to represent various spiking neuron characteristics with appropriate spike response and refractory kernels.

\subsection{Feedforward Spiking Neural Networks}
In this section, we define the model of feedforward SNNs and describe how events and spikes are related.

One of the advantages of SNNs over ANNs is their ability to process event-data from \evcams~directly. \Evcams~have independent sensors at each pixel that respond asynchronously to brightness changes. An event can be described by a tuple $(x, y, t, p)$, where x and y are the location of the pixel from which the event was triggered at time $t$. The polarity $p$ is a binary variable that indicates whether the change in brightness is either positive or negative.
The SNN model in this work has two inputs (i.e. two channels) for each pixel location to account for the polarity of the events. When an event is fed as an input to the network, we refer to it as \emph{spike}. A sequence of spikes is called \emph{spike train} and is defined as $s(t) = \sum_{t^{(f)}\in\set{F}} \delta(t - t^{(f)})$, where $\set F$ is the set of times of the individual spikes.

Our SNN model is a feedforward SNN with $n_l$ layers. In the following definition, $\mat W^{(l)}$ are the synaptic weights corresponding to layer $l$ and $\vct s_\text{in}(t)$ refers to the spikes of the input layer:
\begin{align}
    \vct s^{(0)}(t) &= \vct s_\text{in}(t) \label{input}\\
    \vct u^{(l+1)}(t) &= \mat W^{(l)} (\varepsilon * \vct s^{(l)})(t) + (\nu * \vct s^{(l+1)})(t) \label{membranePotential}\\
    \vct s^{(l)}(t) &= \sum_{\mathclap{t^{(f)}\in\{t|\vct u^{(l)}(t)=\vartheta\}}} \delta(t-t^{(f)}) \label{spikeFunction}\\
    \vct \omega(t) &= \mat W^{(n_l)}(\varepsilon * \vct s^{(n_l)})(t) \label{eq:prediciton}
\end{align}
where $\vct \omega$ is the prediction of the angular velocity. We use the following form of spike response kernel and refractory kernel:

\begin{align}
    \varepsilon(t) &= \frac{t}{\tau_s} e^{1-\frac{t}{\tau_s}} \mathcal{H}(t) \label{spikeResponseKernel}\\
    \nu(t) &= -2\,\vartheta e^{-\frac{t}{\tau_r}}  \mathcal{H}(t) \label{refractoryResponseKernel}
\end{align}
$\mathcal{H}(\cdot)$ is the Heaviside step function; $\tau_s$ and $\tau_r$ are the time constants of spike response kernel and refractory kernel respectively.

Note how the spike response kernel distributes the effect of input spikes over time (eqs. \eqref{membranePotential} \& \eqref{spikeResponseKernel}), peaking some time later and exponentially decaying after the peak. This temporal distribution allows interaction between two input spikes that are within the effective temporal range of the spike response kernel, thereby allowing short term memory mechanism in an SNN. It is pivotal in allowing the network to estimate the sensor's movement and enables prediction of angular velocity.

\subsection{Network Architecture}\label{sec:network_arch}
Our network architecture is a convolutional spiking neural network losely inspired by state-of-the-art architectures for self-supervised ego-motion prediction \cite{gordon2019depth}. It consists of five convolutional layers followed by a pooling and fully connected layer to predict angular velocities. The first 4 convolutional layers perform spatial downsampling with stride 2. At the same time, the number of channels is doubled with each layer starting with 16 channels in the first layer. Table \ref{tab:snn_architecture_hyperparams} shows these layer-wise hyperparameters in more detail. 
It can be seen that there is another set of hyperparameters that are time constants concerned with the decay rate of the spike response and refractory kernels in equation \eqref{spikeResponseKernel} and \eqref{refractoryResponseKernel}. These time constants are increasing with network depth to account for both high event rate from the \evcam~at the input and slower dynamics at the output for consistent predictions.
Table \ref{tab:snn_architecture_hyperparams} lists these layer-wise hyperparameters of the architecture in more detail.

\subsubsection{Global Average Spike Pooling (GASP)}\label{sec:gasp}
So far, the discussed elements of our architecture are regularly encountered in literature of both spiking and artificial neural networks. In recent years, global average pooling \cite{Lin2013NetworkIN} has become prevalent in modern network architectures \cite{he2016deep, Simonyan15} due to their regularization effect. We adapt this line of work for spiking neural networks and introduce global average spike pooling after the last convolutional layer.

To describe GASP, we define a spatial spike-train $\vct S_i(t, x, y)$ resulting from the $i$-th channel of the previous layer as
\begin{align}
    \vct S_i(t, x, y) = \sum_{t^{(f)}\in\set F_i(x, y)} \delta(t - t^{(f)}),
\end{align}
where $\set F_i(x, y)$ is the set of spike times in the $i$-th channel at the spatial location $(x, y)$. Let $g_i(t)$ be the $i$-th output of the pooling operation, then
\begin{align}
    g_i(t) = \sum_{\substack{x\in\{0,\ldots, W-1\} \\ y\in\{0,\ldots, H-1\}}}\vct S_i(t, x, y)\label{eq:gasp_out},
\end{align}
where $W$ and $H$ are width and height of the previous channel. Successive synapses connected to the spike-train $g_i(t)$ are then scaled by $\nicefrac{1}{W\cdot H}$ to introduce invariance with respect to the spatial resolution.

After the spike-train pooling, a fully connected layer connects the spike-trains to three, non-spiking, output neurons for regressing the angular velocity continuously in time. To summarize the computation after the pooling layer, we can reformulate the angular velocity prediction in equation \eqref{eq:prediciton} as
\begin{align}
    \vct \omega(t) =  \frac{1}{N}\left(\varepsilon *\vct W \left[g_1,\ldots, g_C\right]^\top\right)(t)
\end{align}
where $N = W\cdot H$ is the number of neurons per channel in the last convolutional layer with $C$ channels.

\begin{table*}[!ht]
    \centering
    \caption{Hyperparameters of the spiking neural network architecture.}
    \label{tab:snn_architecture_hyperparams}
    \small
    \begin{tabular}{l|llllll}
         Layer-type &  Conv 1 & Conv 2 & Conv 3 & Conv 4 & Conv 5 & Fully connected\\\midrule
         Kernel size & $3\times3$ & $3\times3$ & $3\times3$ & $3\times3$ & $3\times3$ & - \\
         Channels & 16 & 32 & 64 & 128 & 256 & - \\
         Stride & 2 & 2 & 2 & 2 & 1 & - \\
         $\tau_s,\;\tau_r\;[\text{ms}]$ & 2, 1 & 2, 1 & 4, 4 & 4, 4 & 4, 4 & 8, -\\
    \end{tabular}
\end{table*}

\subsection{Synthetic Dataset Generation}\label{sec:data_generation}
Supervised learning of spiking neural networks requires a large amount of data. In our case, we seek a dataset that contains events from an \evcam~with ground truth angular velocity. The three main criteria of our dataset are the following: First, there must be a large variety of scenes to avoid overfitting to specific visual patterns. Second, the dataset must be balanced with respect to the distribution of angular velocities. Third, precise ground truth at high temporal resolution is required. To the best of our knowledge, such a dataset is currently not available. As a consequence, we generate our own dataset.

To fulfill all three criteria, we generated a synthetic datasets using ESIM \cite{Rebecq18corl} as an \evcam~simulator. ESIM renders images along a trajectory and interpolates a brightness signal to yield an approximation of the intensity per pixel at all times. This signal is then used to generate events with a user-chosen contrast threshold. We selected the contrast threshold to be normally distributed with mean $0.45$ and standard deviation of $0.05$. Furthermore, we set the resolution to $240\times 180$ to match the resolution of the DAVIS240C \evcam \cite{brandli2014240}.

As a next step, we selected a subset of 10000 panorama images of the Sun360 dataset \cite{xiao2012recognizing}. From these images, ESIM simulated sequences with a temporal window of 500 milliseconds each. This amounts to approximately 1.4 hours of simulated data. The random rotational motion used to generate this data was designed to cover all axes equally such that, over the whole dataset, angular velocities are uncorrelated and their mean is zero.%

Finally, the dataset is divided into 9000 sequences for training and 500 sequences each for validation and testing.

\subsection{Loss Function}
The loss function $L$ is defined as the time-integral over the euclidean distance between the predicted angular velocity $\vct\omega(t)$ and ground truth angular velocity $\hat{\vct \omega}(t)$:
\begin{align}
    L = \frac{1}{T_1-T_0}\int_{T_0}^{T_1} \sqrt{\vct e(t)^\top \vct e(t)} \diff t\label{eq:loss}
\end{align}
where $\vct e(t) = \vct \omega(t) - \hat{\vct \omega}(t)$. The error function is not immediately evaluated at the beginning of the simulation because the SNN has a certain settling time due to its dynamics. 
Note that this loss function is closely related to the van Rossum distance \cite{rossum2001novel} which has been used for measuring distances between spike-trains.

\subsection{Training Procedure}\label{sec:training_proc}
SNNs are continuous-time dynamical system and, as such, must be discretized for  simulation on GPUs. In the ideal case, we choose the discretization time steps as small as possible for accurate simulation. In practice, however, the step size is a trade-off between accuracy of the simulation and availability of memory and computational resources. We chose to restrict the simulation time to 100 milliseconds with a time step of one millisecond. The loss is then evaluated from 50 milliseconds onwards to avoid punishing settling time with less than 50 milliseconds duration.

The training of our SNNs is based on first-order optimization methods. As a consequence, we must compute gradients of the loss function with respect to the parameters of the SNN. This is done with the publicly available\footnote{\url{https://github.com/bamsumit/slayerPytorch}} PyTorch implementation of SLAYER \cite{shrestha2018slayer}.

We augment the training data by performing random horizontal and vertical flipping and inversion of time to mitigate overfitting. The networks are then trained on the full resolution ($240\times 180$) of the dataset for 240,000 iterations and batch size of 16. The optimization method of choice is ADAM \cite{Kingma2014AdamAM} with a learning rate of $0.0001$ without weight decay.

\section{Experiments}\label{sec:experiments}
In this section, we assess the performance of our method on the dataset described in section \ref{sec:data_generation} to investigate the following questions:
\begin{itemize}
    \item What is the relation between angular velocity and prediction accuracy?
    \item Are the predictions for tilt, pan and roll rates of comparable accuracy?
    \item Is our method competitive with respect to artificial neural networks?
\end{itemize}

\begin{figure*}[ht]
    \centering
    \subfloat[\label{fig:qual_plot:a}]{\includegraphics[width=0.48\linewidth]{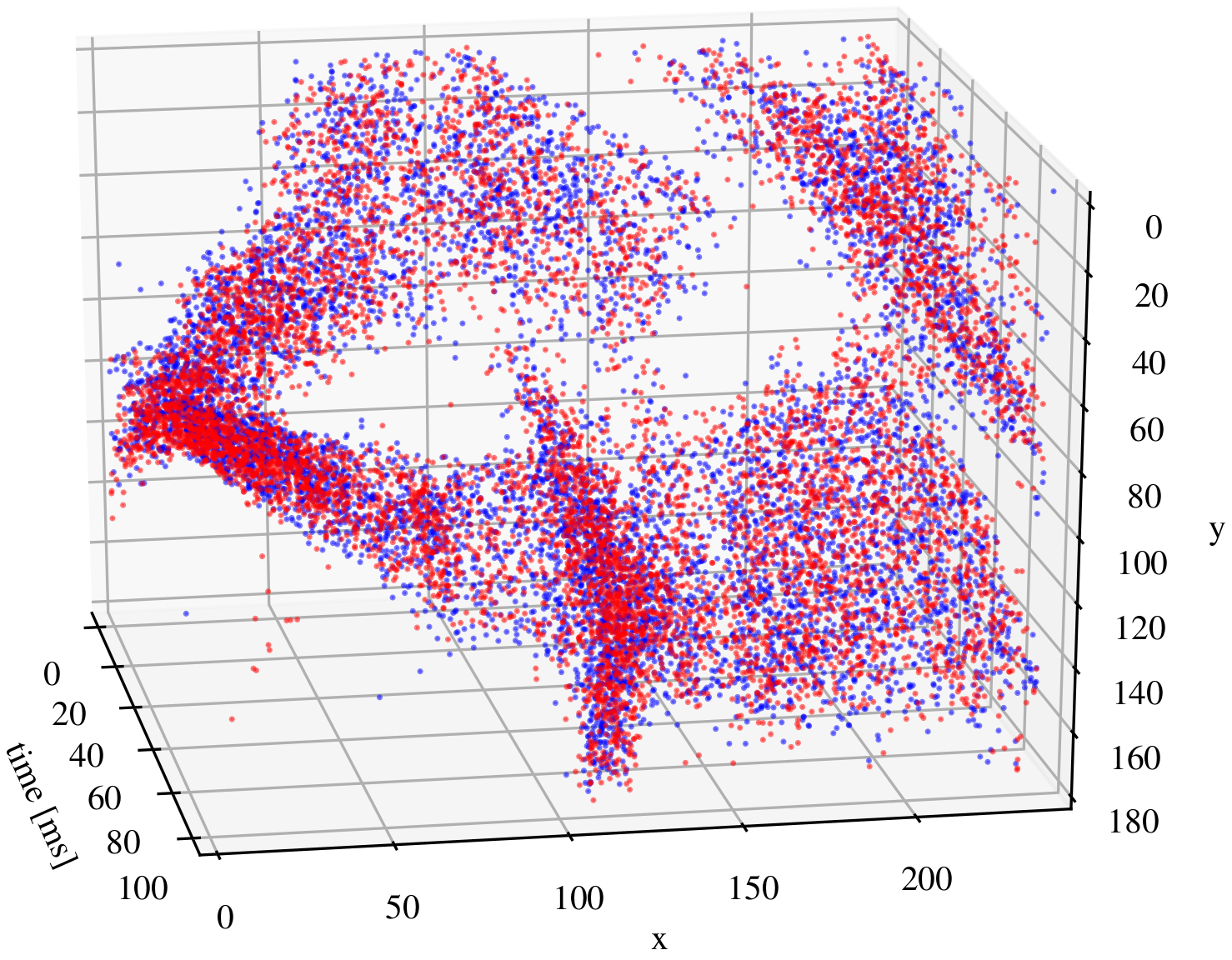}}
    \subfloat[\label{fig:qual_plot:b}]{\includegraphics[width=0.5\linewidth]{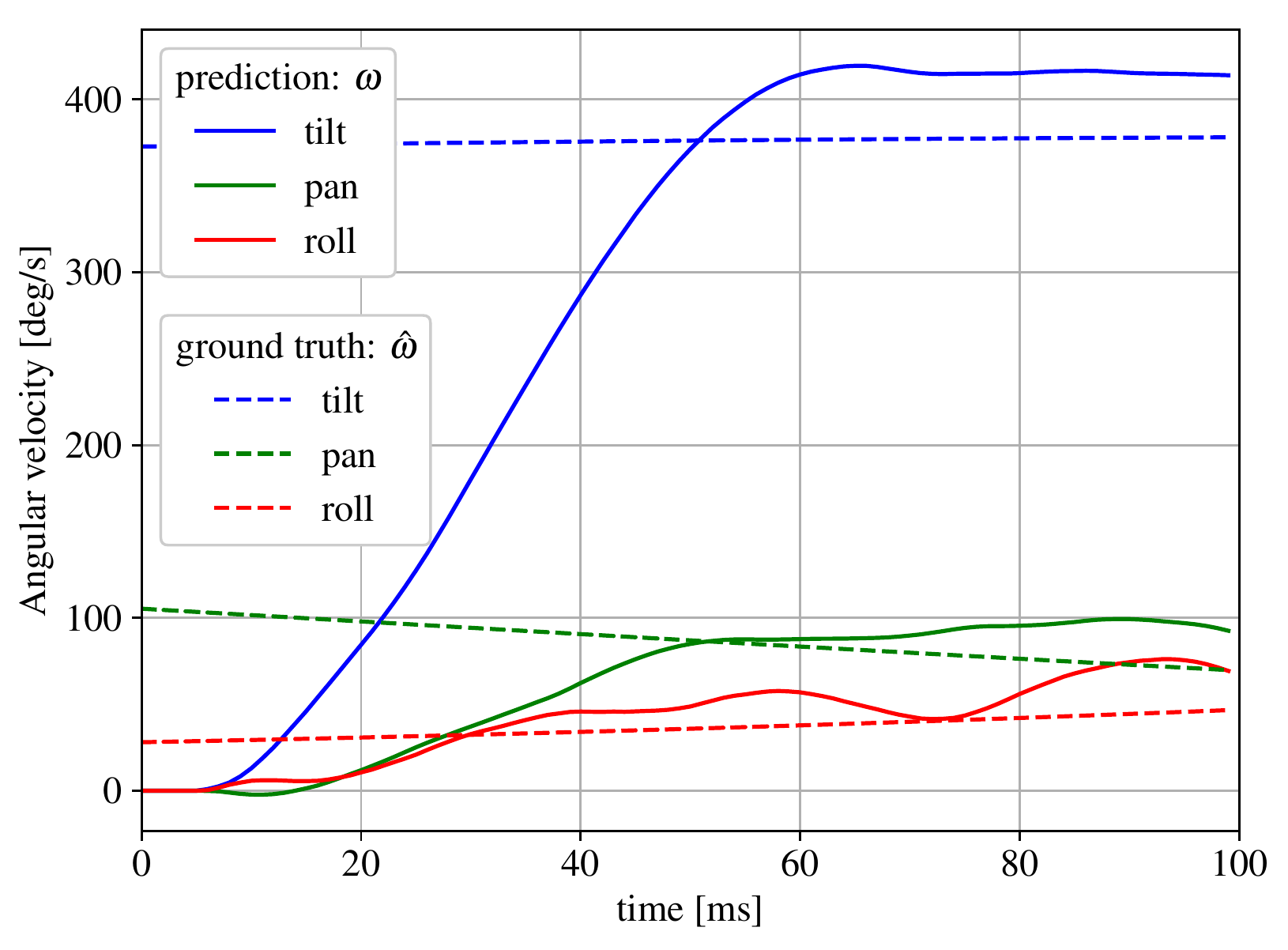}}
    \caption{(a): Events over the 100ms test sequence. Positive events in red and negative events in blue. (b) Continuous-time angular velocity predictions by the SNN and the corresponding ground truth. The SNN requires a settling time of around 50 milliseconds which is exactly when the loss function is applied while training the network. `Pred' refers to prediction and `gt' refers to ground truth.}
    \label{fig:qual_plot}
\end{figure*}

\subsection{Experimental Setup}\label{sec:exp_setup}
For the purpose of evaluating the prediction accuracy of different methods, we split the test set into 6 subsets each containing a specific range of angular velocities. The test set itself is generated in identical fashion to the training and validation set. More importantly, the panorama images, from which the event data of the test set is generated, are unique to the test set.
What makes this dataset especially challenging is the fact that the angular velocity is not initialized at zero but rather at a randomly generated initial velocity. The angular velocity slightly varies within the generated sequence but does not change drastically.

The SNN is simulated with a step size of 1 milliseconds but is only evaluated after 50 milliseconds to eliminate the influence of the settling time on the evaluation accuracy. This is in accordance to the training methodology discussed in section \ref{sec:training_proc}. Figure \ref{fig:qual_plot:b} visualizes the prediction of the SNN over the 100ms sequence. As expected, the network only achieves good tracking after the settling time since it is not penalized for inaccuracies during settling time. Figure \ref{fig:qual_plot:a} also shows the space-time volume of events that are fed to the SNN for the same sequence.

We also compare our method against three feedforward artificial neural networks. Architecture ANN-6 is based on the same architecture as the 6-layer SNN (SNN-6) specified in table \ref{tab:snn_architecture_hyperparams} (with ReLU activation functions). To examine the importance of deeper networks we train two ResNet-50 architectures \cite{he2016deep}. The only difference between them is that one is trained with inputs consisting of two-channel frames computed by accumulating events \cite{Maqueda18cvpr}, denoted by (a), while the other is trained with inputs computed by drawing events into a voxel-grid \cite{Gehrig19arxiv}, denoted by (V). ANN-6 is only trained with the voxel-grid representation.

Feedforward ANNs cannot continuously predict angular velocity\footnote{It is theoretically possible to shift the time window for very small increments at the expense of computational costs}. As a consequence, the training of the ANNs is based on minimizing the error of the mean angular velocity within a time window of 20 ms. Subsequently, the time window is shifted to the next non-overlapping sequence of events. In a similar fashion, ANN predictions are evaluated every 20 ms for comparison with the SNN.

\subsection{Quantitative Evaluation}\label{sec:quant_eval}

\begin{figure}[ht]
    \centering
    \includegraphics[width=0.9\linewidth]{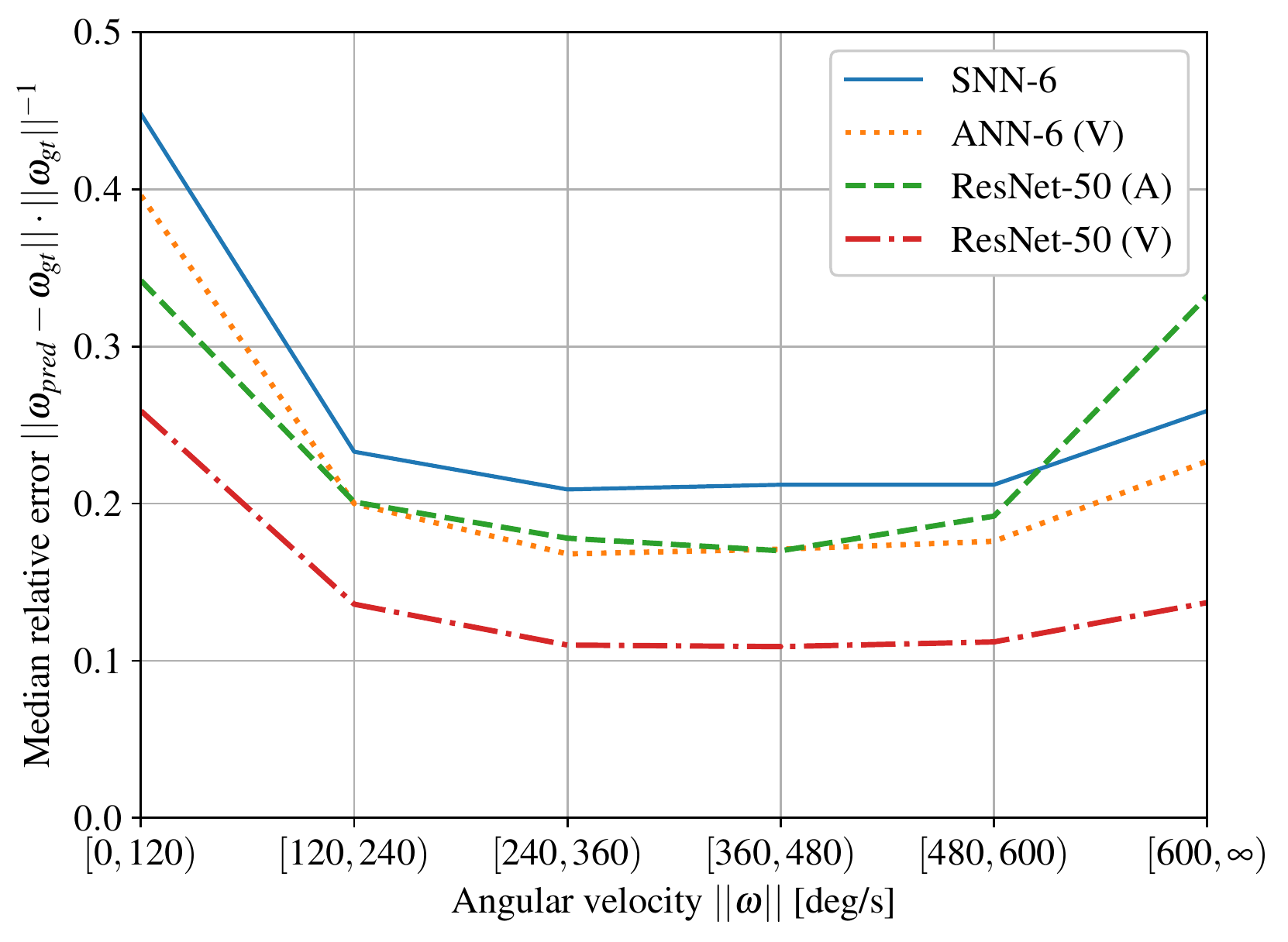}
    \caption{Median relative errors on the test set for different angular velocities for all trained models. $[\omega_a, \omega_b)$ indicates that angular velocities in the range of $\omega_a$ and $\omega_b$ are considered. The SNN achieves comparable accuracy to its ANN counterpart with 6 layers. Both are outperformed by ResNet-50 with the voxel-based input representation (V). In contrast, the same network with accumulation-based input (A) achieves errors on the order of ANN-6 and SNN-6. This highlights that the lack of accurate input representation cannot be compensated with increasing the number of layers in the network.}
    \label{fig:total_rmse}
\end{figure}

Figure \ref{fig:total_rmse} reports the median of the relative error over the range of angular velocities in the test set for all trained models. All models tend to have high relative error at slow angular velocity.
Note, however, that achieving low relative error at low absolute speed is difficult in general due to the fact that the relative error is infinite in the limit of zero angular velocity. Overall, the 6-layer SNN performs comparably to the ANN-6 and the ResNet-50 (A) baseline while ResNet-50 (V) with the voxel-based representation achieves the lowest error in general. These findings are condensed in table \ref{tab:rmse_total} which additionally provides the RMSE and median of  relative errors of the naive mean\footnote{Arithmetic mean of the training dataset which is close to zero} prediction baseline.

\begin{figure}[ht]
    \centering
    \includegraphics[clip, trim=15pt 15pt 15pt 15pt, width=1.0\linewidth]{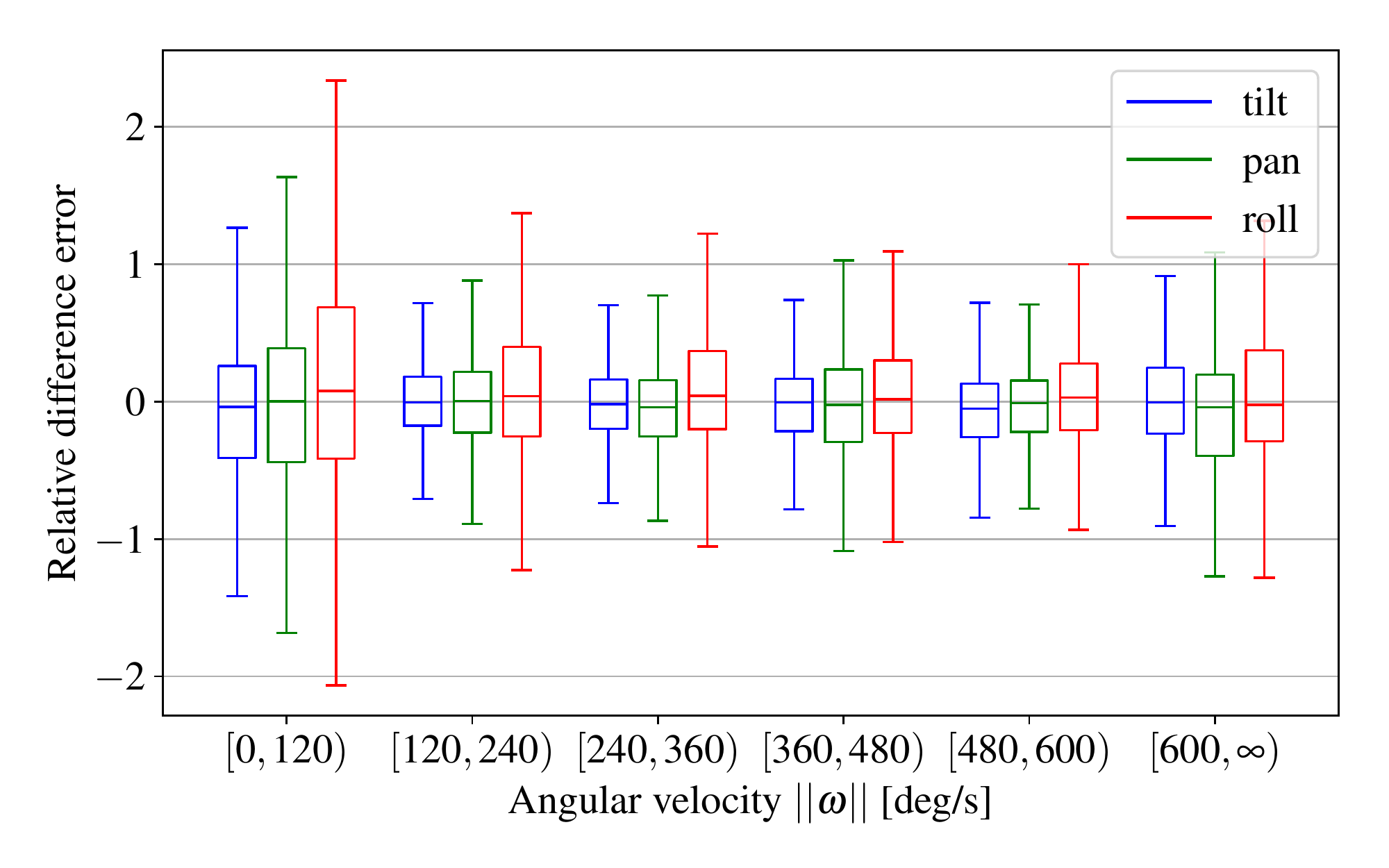}
    \caption{Quartiles of the relative difference errors of SNN predictions on the test set. The difference between prediction and groundtruth is normalized with respect to the absolute value of the ground truth tilt, pan or roll rates respectively. Evidently, the SNN is performing better at moderate to high angular rates while the roll predictions are in general less accurate than tilt and pan.}
    \label{fig:rel_diff_error_snn}
\end{figure}

Next, we investigate the impact of angular velocities on tilt, pan and roll rates separately. Figure \ref{fig:rel_diff_error_snn} shows the box plots of the relative errors\footnote{defined as $\frac{\omega_{i} - \hat\omega_{i}}{|\hat\omega_{i}|}$, with $i$ for either the tilt, pan or roll axis} with respect to different angular velocities. Across the whole range of angular velocities, predictions for tilt are slightly more accurate than those for pan while the error for roll is in general higher than compared to the other axes.

\subsection{Discussion}
In summary, the SNN is able to regress angular velocity with reasonable accuracy across different rates. The 6-layer ANN achieves only slightly lower error than the SNN. %
From this result we conclude that it is possible to train SSNs to ANN-competitive accuracy on this continuous-time regression task. The slightly lower performance could originate from potentially suboptimal hyperparameters for spike and refractory response ($\tau_s$ and $\tau_r$ in table \ref{tab:snn_architecture_hyperparams}). These parameters could potentially be learned as well but this is left to future work.

The large discrepancy between the error achieved by the two ResNet architectures are due to their difference in the input representation. Unlike the voxel-based representation, the accumulation-based representation completely discards timings of the events. This appears to be problematic for regression of angular velocity.
On the other hand, the significant jump in accuracy from ANN-6 to ResNet-50, both with voxel-based input, suggests that the SNN could also benefit from increasing the number of layers. Nevertheless, we expect that optimizing deeper SNNs might uncover new challenges for currently popular training methods \cite{shrestha2018slayer, Neftci2019}.%

Our axis-isolating experiments suggest that predicting roll rate is more challenging for the SNN than predicting tilt and pan rates. Similar observations were made for an optimization-based approach to angular rate tracking \cite{Gallego17ral}. When the camera is being rolled, events are typically triggered at the periphery. The resulting spatial-temporal patterns are spread over the whole frame, which poses difficulties for our architecture.

\begin{table}[ht]
    \centering
    \footnotesize
    \caption{Baseline comparisons on the test set: The SNN is compared against the ANN models and the naive mean prediction baseline. Input representations are either event-based (E), accumulation-based (A) \cite{Maqueda18cvpr} or voxel-based (V) \cite{Gehrig19arxiv}.%
    }
    \label{tab:rmse_total}
    \begin{tabular}{llllll}
        & mean & SNN-6 & ANN-6 & \multicolumn{2}{l}{ResNet-50}\\\toprule
        Relative error & $1.00$ & $0.26$ & $0.22$ & $0.22$ & $0.15$\\
        RMSE (deg/s) & $226.9$ & $66.3$ & $59.0$ & $66.8$ & $36.8$\\
        Input type & - & E & V & A & V\\
    \end{tabular}
\end{table}

\section{Conclusion}\label{sec:conclusion}
In this work, we investigated the applicability of feedforward SNNs to regress angular velocities in continuous-time. We showed that it is possible to train a spiking neural network to perform this task on par with artificial neural networks. Thus, we can confirm that state-of-the-art SNN training procedures accurately address the temporal error assignment problem for SNNs of the size as presented in this work. Experimental results further suggest that deeper SNNs might perform significantly better, but there are a number of obstacles ahead. Backpropagation-based approaches require that we unroll the SNN in time at high-resolution. This requirement poses serious challenges for optimization on GPUs both in terms of memory consumption and FLOPS. It has been a long-standing research goal to address these issues and we believe it to be crucial to unlock the full potential of SNNs.

\todo{Add qualitative example, e.g. event-frame, with angular velocity for illustration}

{\small
\bibliographystyle{ieeetr}
\bibliography{root}
}

\end{document}